\documentclass[a4paper, 10pt, conference]{ieeeconf}      

\IEEEoverridecommandlockouts                              

\usepackage{graphics} 
\usepackage{epsfig} 
\usepackage{mathptmx} 
\usepackage{times} 
\usepackage{amsmath} 
\usepackage{amssymb}  
\usepackage{algorithm}
\usepackage{algorithmic}
\usepackage{epstopdf}
\AppendGraphicsExtensions{.pdf}
\usepackage{makecell}
\usepackage{tabularx}
\usepackage{graphicx}
\usepackage{amssymb}
\usepackage{amssymb}

\title{\LARGE \bf
 Object Detection for Understanding Assembly Instruction Using Context-aware Data Augmentation and Cascade Mask R-CNN
}

\author{Joosoon Lee$^{1}$, Seongju Lee$^{1}$, Seunghyeok Back$^{1}$, Sungho Shin$^{1}$, and Kyoobin Lee$^{1}$
\thanks{$^{1}$J. Lee, S. Lee, S. Back, S. Shin, and K. Lee are with the School of Integrated Technology (SIT), Gwangju Institute of Science and Technology (GIST), Cheomdan-gwagiro 123, Buk-gu, Gwangju 61005, Republic of Korea.
        {\tt\small kyoobinlee@gist.ac.kr}}%
}

\begin{document}
\maketitle
\thispagestyle{empty}
\pagestyle{empty}

\begin{abstract}

Understanding assembly instruction has the potential to enhance the robot's task planning ability and enables advanced robotic applications. To recognize the key components from the 2D assembly instruction image, We mainly focus on segmenting the speech bubble area, which contains lots of information about instructions. For this, We applied Cascade Mask R-CNN and developed a context-aware data augmentation scheme for speech bubble segmentation, which randomly combines images cuts by considering the context of assembly instructions. We showed that the proposed augmentation scheme achieves a better segmentation performance compared to the existing augmentation algorithm by increasing the diversity of trainable data while considering the distribution of components locations. Also, we showed that deep learning can be useful to understand assembly instruction by detecting the essential objects in the assembly instruction, such as tools and parts.


\end{abstract}

\section{INTRODUCTION}

Task planning from implicit instruction is crucial for robots to perform complex tasks in the real world. Since the user’s instructions for the robot can be abstract and ambiguous, a robot is required to set an explicit objective then divide it into sub-goals and perform proper skills while checking the current status and planning the next strategy. However, the current level of robots requires a human to manually give them a detailed direction with pre-defined planning and specific conditions, while human learns a new task by understanding an instruction, such as work statement, operation manual, and assembly instruction. As those instruction documents can be easily found in many areas, intelligence that can understand instructions visually and establish a task plan has the potential to automate various industrial processes and manufacturing systems in robotics.

IKEA assembly instructions are one of the typical and challenging instruction types, expressed in a symbolic, high-dimensional, unstructured way. Most IKEA instructions are multi-page and composed of texts, symbols, and diagrams, and objects appear with various viewpoints, sizes, and occlusion s. In addition, it depicts only a few instances of the whole process, so detail instructions are omitted. Thus, to generate a task plan from assembly instruction, the algorithm should infer the explicit process from the symbolic 2D image while considering the relations between multiple components and steps.


\begin{figure}[t!]
\centering
 \includegraphics[width=0.48\textwidth]{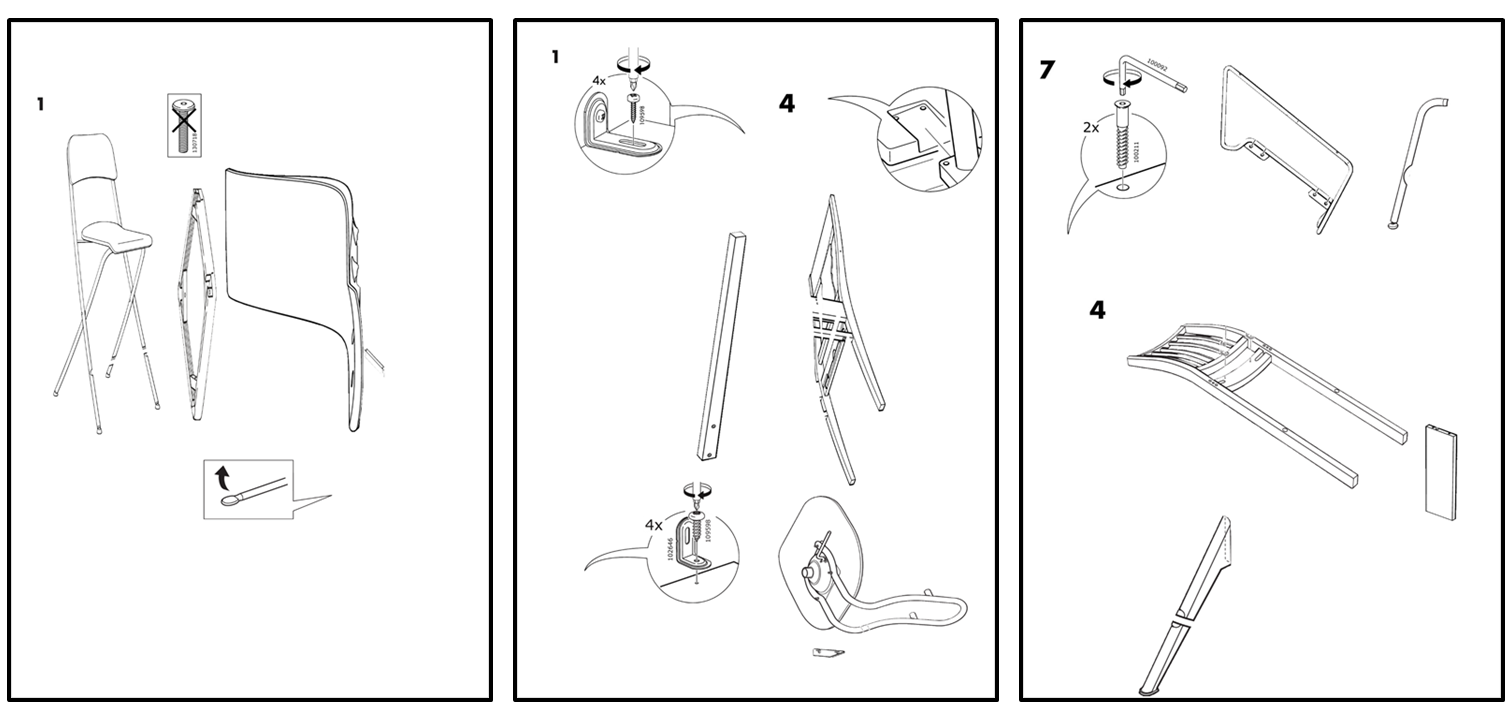}
 \caption{Example images of augmented assembly instructions using our context-aware augmentation}
 \label{fig:aug-example}
\end{figure}

In furniture assembly instruction, there are several components in each step, such as furniture parts, mechanical connectors, symbols, and numbers \cite{shao2016dynamic}. Among these elements, understand the contents in speech bubbles is crucial since they contain specific assembly directions by highlighting relatively small connection parts in assembly instructions. The speech bubble contains the detailed instructions about what connectors should be used with which tools and actions. Therefore, the segmentation of speech bubbles is essential for the automation of analyzing instructions.

Shao et al.\cite{shao2016dynamic} proposed a method that segments and recognizes the visual elements in IKEA assembly instructions using supervised learning. They showed that reconstruction of 3D models from assembly instructions in a vector graphic could be accomplished by using the HO feature and random regression forest \cite{fanelli2011real}. However, their system requires an elaborate pre-processing and parsing step to segment visual elements. User interaction is also necessary due to the lack of generality over various manuals and furniture types. 

Towards a fully automated system that can generate a task plan from an assembly instruction, deep learning can be utilized to detect and analyze the key components from the instructions. Deep learning-based object detection and segmentation algorithms, which show the state-of-the-art performance \cite{ren2015faster, he2017mask, redmon2016you, zhang2018single, cai2018cascade}, could recognize and localize objects of various sizes, shapes, and locations from a 2D image. However, a deep learning model requires a large-scale dataset for accurate performance, while data labeling is tedious and costly labor. Thus, data augmentation, which adds a variation on the training set, can be useful in improving detection performance and preventing over-fitting \cite{shorten2019survey, taylor2018improving, cubuk2019autoaugment, bagherinezhad2018label}.

In this study, we present a deep learning-based method to understand the assembly process depicted in assembly instruction. We apply Cascade Mask R-CNN \cite{cai2018cascade} to segment speech bubbles and detect key components in instructions comprising a 2D image. In addition, we propose a context-aware data augmentation methods to reduce the labeling cost and improve the performance of speech bubble segmentation. Through experiments, we showed that the proposed method could extract essential information for generating a task plan from assembly instructions with a data-driven approach.
\newline





\section{Method}
\subsection{IKEA Assembly Instruction Annotation}
\subsubsection{Dataset Statistics}
To construct a dataset for assembly instructions, unstructured image data, we collected and annotated images from IKEA which is one of the most popular assembly furniture. Assembly instructions of 193 kinds furniture contained 2,673 images. The common sizes of instruction images are 2,339 $\times$ 1,654 and 1,166 $\times$ 1654, and the images are grayscale images. The objects we annotated are speech bubble, part, tool, symbol, and text. The statistics of each object are listed in Table \ref{tb:statistics}.

\begin{table}[h]
\caption{Annotation Statistics according to the Instance Type}
\label{tb:statistics}
\begin{center}
\begin{tabular}{cccc}
\Xhline{2\arrayrulewidth}
\textbf{Instance} & \textbf{Annotation} & \textbf{Number of} & \textbf{Number of} \\ \textbf{Name} &  \textbf{Type}& \textbf{Instances} & \textbf{Images} \\
\hline
Speech bubble & Polygon & 393 & 318 \\

Part & BBOX & 896 & 435 \\

Tool & BBOX & 1,589 & 1,081 \\

Symbol & BBOX & 794 & 326 \\

Text & BBOX & 2,518 & 503 \\
\Xhline{2\arrayrulewidth}
\end{tabular}
\end{center}
\end{table}

\subsubsection{Annotation Attributes}
We annotated category and location of each object for detection and segmentation. For segmentation, pixels of each instance were masked corresponding category value with a polygon. For example, the background pixels were assigned 0, and each pixel of objects was assigned value to corresponding each category. For detection, bounding boxes were used which can be expressed with width, height, and two values for 2D coordinate of center point of object in an image.  

In the process of dataset construction, five kinds of categories were annotated. The categories are speech bubble, part, tool, symbol, and text. Speech bubbles are annotated for segmentation. Speech bubble usually consists of body and tail (there are some speech bubbles which have no tails). Circle shaped bodies and winding tails are annotated with same manner of pixel-wise masking. Some part pixels popped out of the body were also annotated with the same value as the body. The objects of part, tool, symbol and text were annotated for detection with bounding box. Parts means screws, which are mainly located in bodies of speech bubbles. Some parts located out of speech bubbles are annotated when the whole head and tail of the parts are expressed. Tools include hand, hammer, driver, wrench, spanner, drill, and awl. Symbols are filled with black and located in the area indicating assembly actions. Text objects are numbers or characters expressing ID of parts and tools or information of instruction pages.
\newline

\begin{figure}[t!]
\centering
    \includegraphics[width=0.48\textwidth]{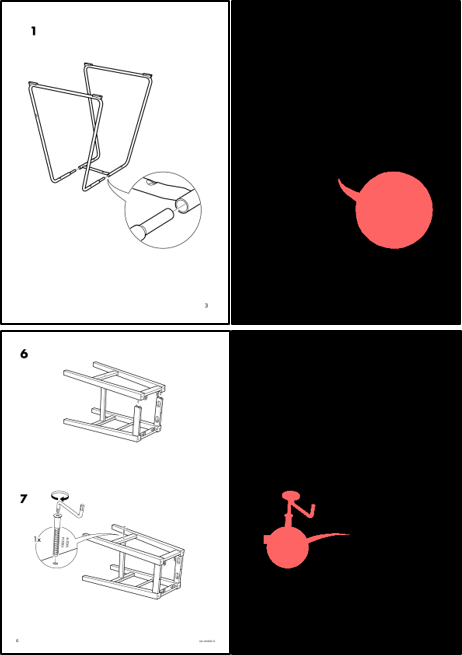}
    \caption{IKEA assembly instruction images and speech bubble mask images}
    \label{fig:ann-example-seg}
\end{figure}

\begin{figure}[t!]
\centering
    \includegraphics[width=0.48\textwidth]{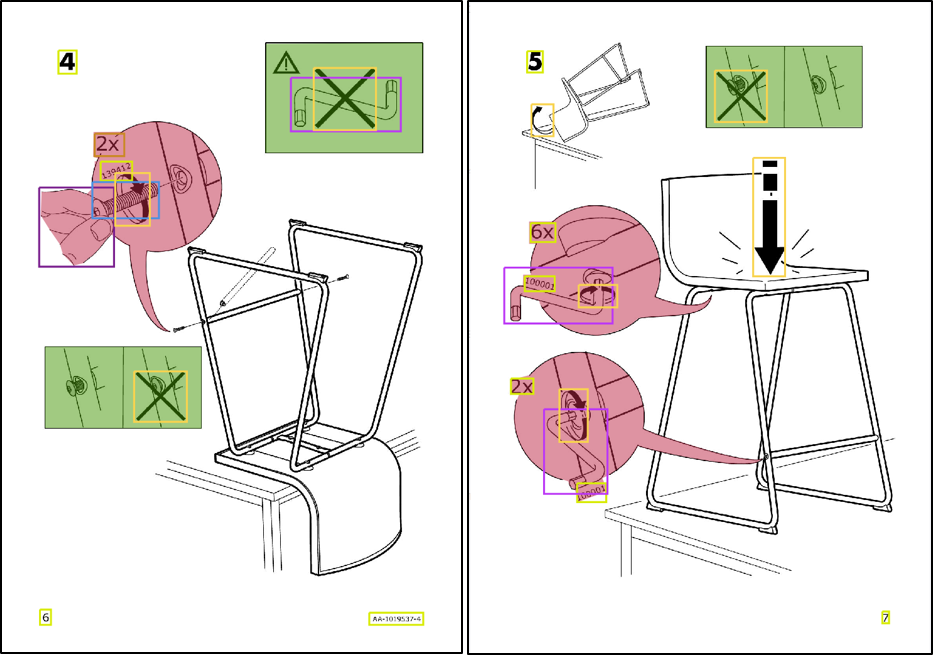}
    \caption{Examples of annotated IKEA assembly instructions}
    \label{fig:ann-example-det}
\end{figure}

\subsection{Context-aware Data Augmentation}
In order to overcome lack of training data in segmentation, we designed a data augmentation algorithm which considers the context of IKEA assembly instruction. We analyzed the characteristics of IKEA assembly instructions such as arrangement of components in instructions and set some rules based on the characteristics to cut and paste key components. Fig. \ref{fig:aug-example} shows examples of the synthesized IKEA assembly instructions generated by our proposed algorithm.

\begin{figure}[t!]
\centering
 \includegraphics[width=0.45\textwidth]{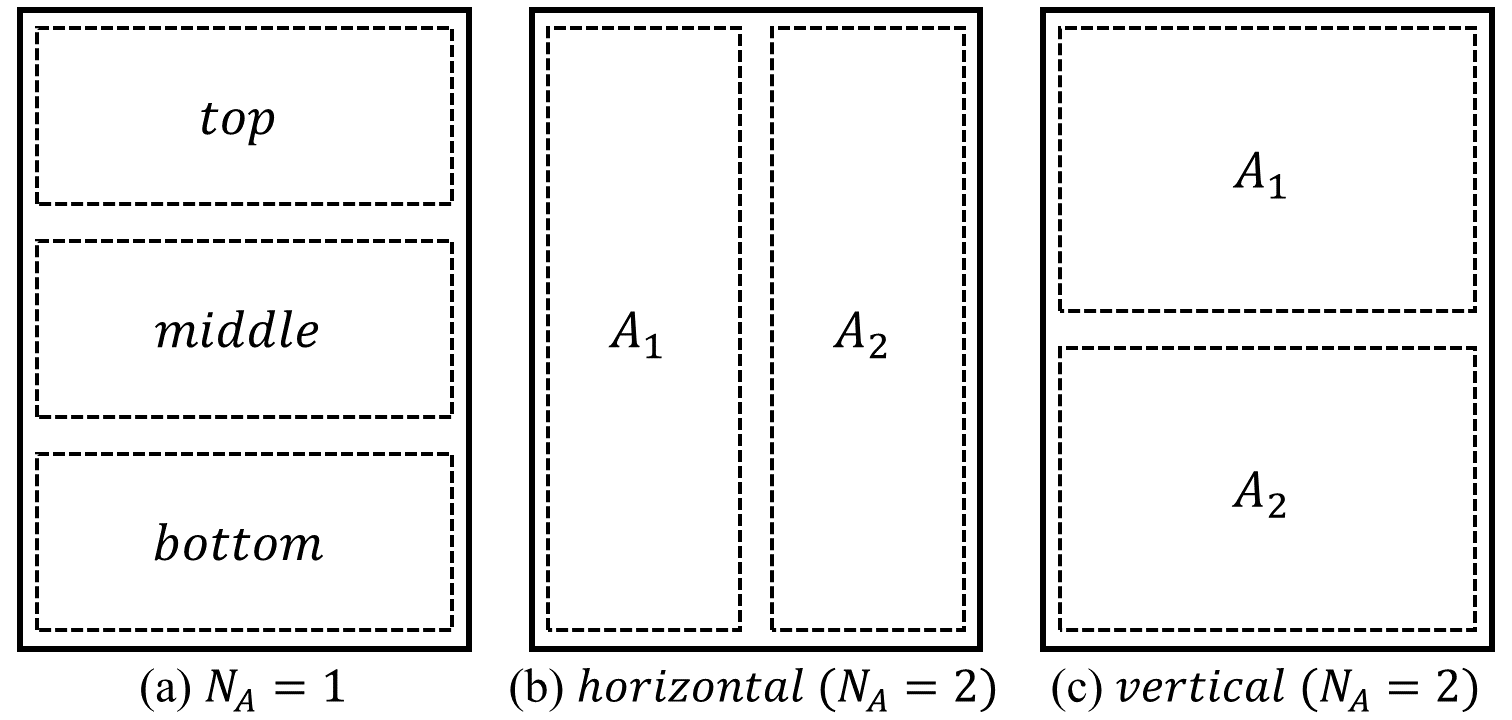}
 \caption{Arrangement of effective area. (a) describes three types arrangement when $N_{A}=1$. (b) and (c) describe horizontal and vertical arrangement when $N_{A}=2$, respectively.}
 \label{fig:arrangement}
\end{figure}

First, we defined effective area ($A$) which refers to the area where one assembly stage can be located. Then the number, arrangement, and size of effective area are determined. The number of effective area ($N_{A}$) is determined as one or two since one page of IKEA assembly instructions mostly contains one or two assembly stage. After determining $N_{A}$, arrangement is chosen as described in line \ref{a1-1} and line \ref{a1-2} of Algorithm \ref{a1}. The diagram of arrangement is described in Fig. \ref{fig:arrangement}. After determining the arrangement of effective areas, $\alpha$ and $\beta$ are randomly chosen in the manually defined range, where $\alpha$ and $\beta$ are reduction ratio of width and height of the document, respectively. Therefore, the width and height of the effective area can be described as the following equations:
\newline
\begin{equation}
    (W_{A}, H_{A})  = (\alpha W, \beta H)
\end{equation}
\noindent
where $W_{A}$ and $H_{A}$ are width and height of effective area, respectively, and $W$ and $H$ are width and height of document, respectively.
\newline

\begin{algorithm}
\caption{Determination of $\alpha$ and $\beta$}
\label{a1}
\textbf{Input}: $N_{A} \in \{1, 2\}$ \newline
\textbf{Output}: $\alpha, \beta \in \mathbb{R}$
\begin{algorithmic}[1]
\IF{$N_{A}==1$}
\STATE{$Arrangement \leftarrow Sampler(\{top, middle, bottom\})$}\label{a1-1}
\STATE{$\alpha_{1} \leftarrow Sampler([0.7, 0.9])$}
\IF{$Arrange==top$ \OR $Arrange==bottom$}
\STATE{$\beta_{1} \leftarrow Sampler([0.4, 0.6])$}
\ELSIF{$Arrange==middle$}
\STATE{$\beta_{1} \leftarrow Sampler([0.6, 0.8])$}
\ENDIF
\ELSIF{$N_{A}==2$}
\STATE{Set $Arrangement \in \{horizontal, vertical\}$}\label{a1-2}
\FOR{each $A_{i}$}
\IF{$Arrange==horizontal$}
\STATE{$\alpha_{i} \leftarrow 0.5$}
\STATE{$\beta_{i} \leftarrow Sampler([0.7, 0.9])$}
\ELSIF{$Arrange==vertical$}
\STATE{$\alpha_{i} \leftarrow Sampler([0.7, 0.9])$}
\STATE{$\beta_{i} \leftarrow Sampler(\{\frac{1}{3}, \frac{1}{2}, \frac{2}{3}\})(\sum_{i}{\beta_{i}} \leq 1)$}
\ENDIF
\ENDFOR
\ENDIF
\end{algorithmic}
$[a, b]$: Closed interval between $a$ and $b$ \newline
$Sampler()$: Random sampler \newline
$\{ \}$: Set 
\end{algorithm}

\begin{figure}[t!]
\centering
    \includegraphics[width=0.48\textwidth]{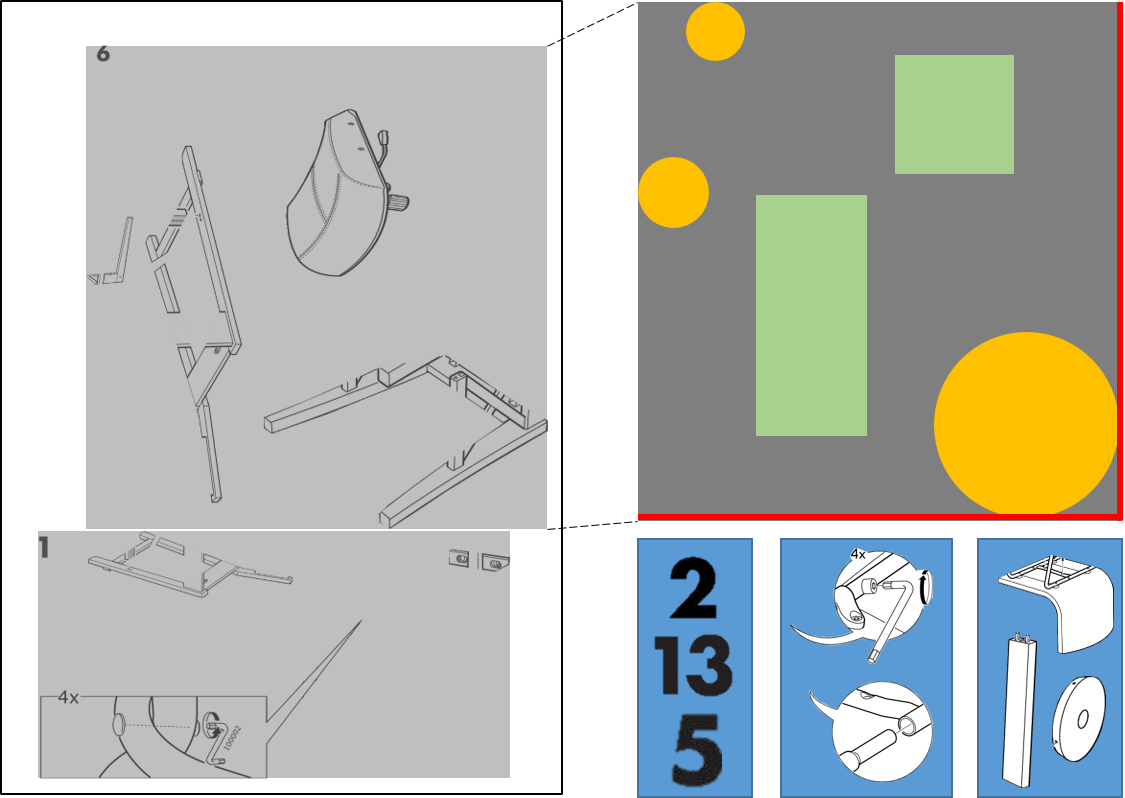}
    \caption{Components of context-aware data augmentation. Gray boxes mean effective areas. In a effective area, yellow circles mean edge components and red lines denote shared edges. Center components are expressed as green boxes. Types of components are number, speech bubble and assembly group, which are in the blue boxes gathered with each category.}
    \label{fig:data-augmentation}
\end{figure}

We defined three types of key components as number, speech bubble, and assembly parts group. After determining $N_{A}$ and size of $A_{i}$, we determine the number and the arrangement of key components located in $A_{i}$. Based on the arrangement characteristics of IKEA assembly instruction, we distinguished key components according to the location: edge and center. Edge components are located at the edge of the effective area, so they meet at least one side of the effective area. On the other hand, we defined center component as a component that does not meet the side of the effective area. Since one effective area (assembly stage) should contain one assembly stage number and one assembly group, the number of edge components should be at least two; it should be four at maximum if components located in all four corners. Then, the type of each edge components is determined (Edge component at top-left corner should be the assembly stage number). For each $A_{i}$, according to the number of components, the arrangement of edge components is determined by determining shared edges. For example, if the number of edge component is three, two edges should contact with the same component (those two edges are shared edges). Then, the remaining procedures are choosing component data from annotated IKEA assembly instructions and pasting them to the augmented instruction. Similarly, the center components are pasted in sequence of determining location and choosing data. The terminologies such as key components and effective area are visualized at Fig. \ref{fig:data-augmentation}.

\subsection{Cascade Mask R-CNN for Segmentation and Detection}
We considered  object recognition in assembly instructions as a visual recognition task in the image data, and we apply a deep learning-based approach. To recognition the body and tail of speech bubble, the object segmentation task, classifying the category of every pixels in image. The objects of tools and parts are detected with bounding boxed generated by deep learning model. The Mask R-CNN model based on two-stage sequence which predicts the location and category of objects is used for detection and segmentation task.  
	
To overcome the limit from inaccurate localization of Mask R-CNN, the cascade training strategy of multi stages method is applied to train deep learning models. We use Cascade Mask R-CNN model for recognizing assembly instructions \cite{cai2018cascade}. Unlike the original Mask R-CNN model, there are additional multiple classifiers in the Cascade Mask R-CNN model. In sequence, the bounding box created by the previous classifier is received as an input of the next classifier. Performance can be maximized by setting a higher standard of intersection of union (IoU) for classifiers.

\section{Experiment}

\subsection{Speech Bubble Segmentation}
In order to apply the deep learning-based method to segment speech bubbles, we used Cascade Mask R-CNN as a baseline network. In speech bubble segmentation, the model was trained by 254 images and evaluated by 64 images. 
Performance indicators of average precision (AP), average recall (AR) and IoU are used to evaluate the performance of speech bubble segmentation through deep learning model, Cascade Mask R-CNN. The results of AP, AR and mean intersection of union (mIoU) are shown at Table \ref{tb:segmentation_result_APAR} as the Original Instruction Data.

\subsection{Data Augmentation for Segmentation}
To check the performance of data augmentation in the segmentation task, we conducted additional training by three types of data augmentation method, \textbf{Naïve Cut-Paste}, \textbf{Instance-Switching}, and \textbf{Context-aware Cut-Paste}. Also, Cascade Mask R-CNN is employed to assess the performance improvement due to our proposed augmentation scheme. Each type of augmented data has 1,000 images, and the models trained by augmented data were evaluated by the 64 images same as original dataset.

\textbf{Naïve Cut-Paste}: Dwibedi \textit{et al.} \cite{Dwibedi_2017_ICCV} presented a simple augmentation method that automatically cut object instances in patch-level and paste them on random background images for supplying sufficient training data to object detectors. In order to implement Naïve Cut-Paste on the assembly instruction, two to eight key components were randomly placed on the white background. Quantities of each key component in one page satisfy the following rules:
\newline
\begin{itemize}
    \item $N_{KeyComponents}$ is randomly sampled from two to eight
    \item $N_{Speechbubble}=min(4, N_{KeyComponents}-2)$
    \item $N_{AssemblyStageNumber}$ and $N_{AssemblyGroup}$ are randomly determined on the condition that the sum of them is $N_{KeyComponents}-N_{Speechbubble}$
    \newline
\end{itemize}

\textbf{Instance-Switching}: Wang \textit{et al.} \cite{wang2019data} recently proposed instance-switching (IS) strategy to solve the problem of existing Cut-Paste methods. This method generated data by simply switching instances of same class from different images and preserved the contextual consistency of the original dataset. Consequently, they improved the performance of object detection on MS COCO benchmark \cite{lin2014microsoft} over various state-of-the-art object detection models. In this work, adopting \cite{wang2019data}, we synthesized virtual assembly instructions by switching speech bubbles from different instruction images to compare the segmentation performance with our method.

\begin{figure}[t!]
\centering
    \includegraphics[width=0.48\textwidth]{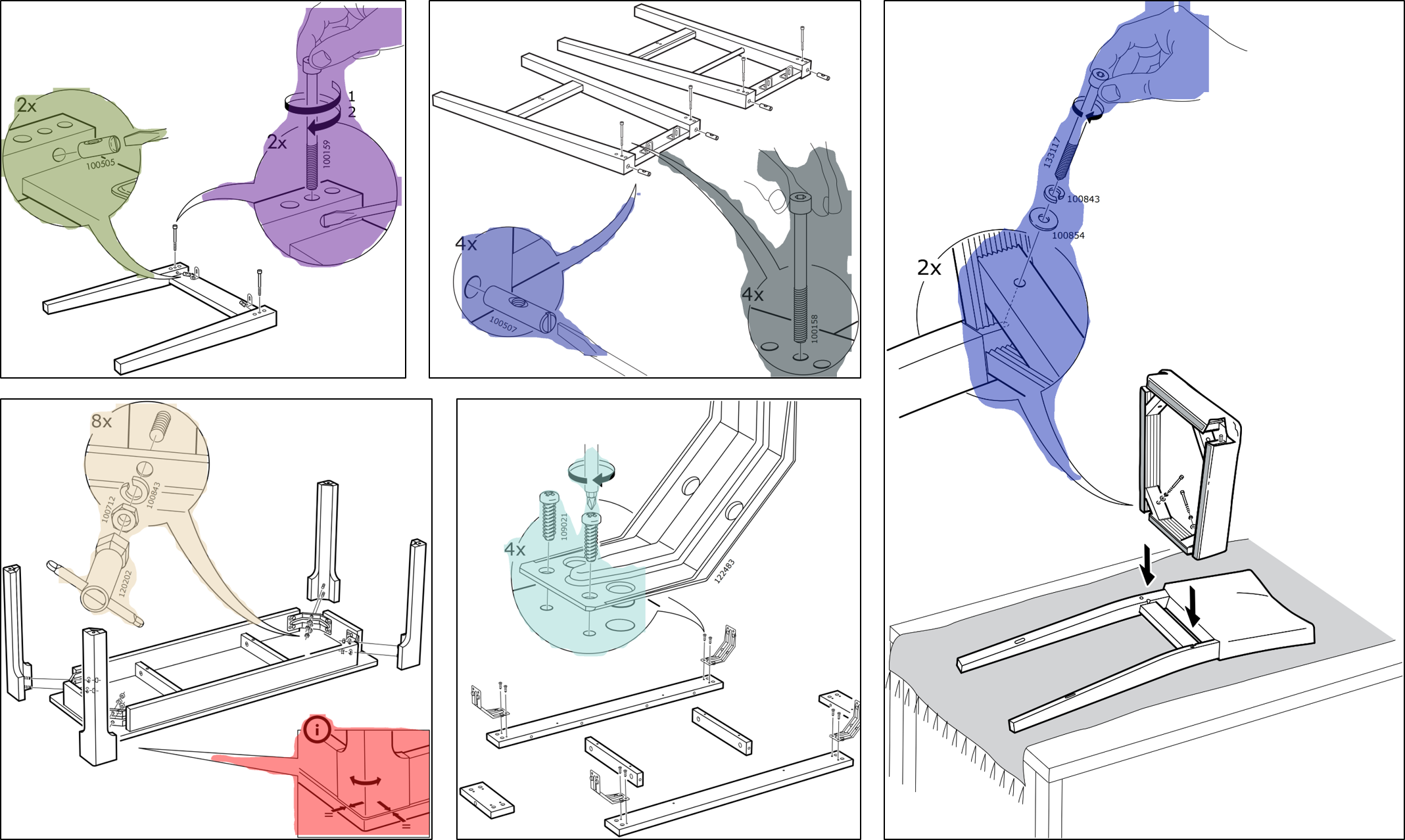}
    \caption{Result examples of speech bubble segmentation visualizing the output of Cascade Mask R-CNN trained Context-aware data augmentation.}
    \label{fig:result-segm}
\end{figure}

We compare the result of context-aware data augmentation with other method. The results of AP, AR and mIoU are shown at Table \ref{tb:segmentation_result_APAR}. The result of context-aware data augmentation reaches the highest performance. We analyze that context-aware augmentation helped the deep learning model to increase generalization by augmenting similarly to the original data distribution. Some examples of visualizations of segmentation output trained by our context-aware data augmentation are shown in Fig. \ref{fig:result-segm}.

\begin{table}[h]
\caption{Results of Object Segmentation in assembly instructions}
\label{tb:segmentation_result_APAR}
\begin{center}
\begin{tabular}{cccccc}
\Xhline{2\arrayrulewidth}
\textbf{Augmentation Method} & \textbf{AP} & \textbf{AP$_{50}$} & \textbf{AP$_{75}$} & \textbf{AR} & \textbf{mIoU}\\
\Xhline{2\arrayrulewidth}
Original Instruction Data & 0.374 & 0.475 & 0.426 & 0.743 & 0.712 \\
\hline
Naïve Paste \cite{Dwibedi_2017_ICCV} & 0.375 & 0.491 & 0.457 & 0.787 & 0.819 \\

Instance Switching \cite{wang2019data} & 0.222 & 0.270 & 0.270 & 0.304 & 0.380 \\

\textbf{(Ours)} Context-aware Paste & \textbf{0.404} & \textbf{0.512} & \textbf{0.488} & \textbf{0.791} & \textbf{0.836} \\
\Xhline{2\arrayrulewidth}
\end{tabular}
\end{center}
\end{table}

\subsection{Object Detection}
In the object detection task, we trained four kinds of objects in the assembly instructions: tool, part, symbol and text. For each category, deep learning models of Cascade Mask R-CNN were trained respectively. In total four deep learning models were trained with 80 percent of annotated images and evaluated with 20 percent of annotated images. 

\begin{figure}[t!]
\centering
    \includegraphics[width=0.48\textwidth]{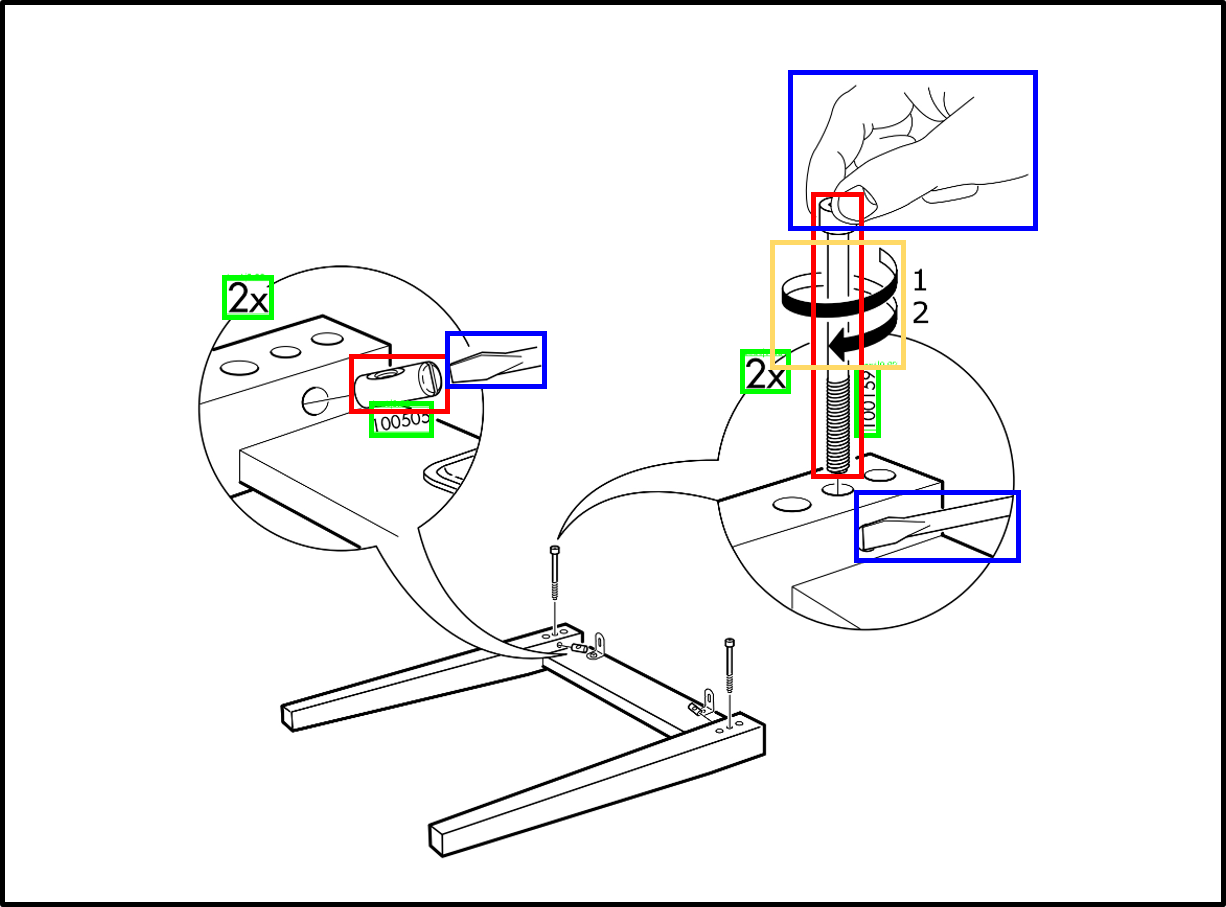}
    \caption{Result example of object detection in assembly instruction. Green boxes contain text objects, red boxes contain parts, blue boxes contain tools, and yellow boxes contain symbols }
    \label{fig:result-det}
\end{figure}

The trained results of average precision and average recall of each object are shown in TABLE \ref{tb:detection_result}, and an example of an output image is shown in Fig. \ref{fig:result-det}. As shown the detection output, text relatively small in size also detected well. It is analyzed that these detection models will be used to identify objects in the assembly instructions and to infer abstract meaning expressed as combinations of objects.

\begin{table}[h]
\caption{Results of Object detection in assembly instructions}
\label{tb:detection_result}
\begin{center}
\begin{tabular}{cccccc}
\Xhline{2\arrayrulewidth}
\textbf{Category} & \textbf{Train Images} & \textbf{AP} & \textbf{AP$_{50}$} & \textbf{AP} & \textbf{AR}\\
\hline

Part & 348 & 0.681 & 0.760 & 0.758 & 0.802 \\

Tool & 864 & 0.733 & 0.835 & 0.811 & 0.808 \\

Symbol & 260 & 0.853 & 0.992 & 0.972 & 0.884 \\

Text & 402 & 0.789 & 0.951 & 0.815 & 0.838 \\
\Xhline{2\arrayrulewidth}
\end{tabular}
\end{center}
\end{table}

\section{CONCLUSIONS}
 Our research focuses on segmenting the speech bubble and detecting the key components, which contain lots of information about instructions, using Cascade Mask R-CNN. However, gathering the training dataset for deep learning is labor-intensive and time-consuming tasks. To ameliorate the aforementioned flaw, we propose context-aware augmentation method, which combines the image cut from the other instructions to increase the number of training data and maximize the advantages of deep learning by overcoming the shortage of manually annotated dataset. The Cascade Mask R-CNN model trained with the augmented data by context-aware method achieves superior segmentation performance compared to the models trained with original IKEA instruction data and augmented data by existing augmentation.  
 
 As a future plan, we will make the advanced model of Mask-RCNN, which could detect the other components such as furniture part, number, and text instructions with the augmentation technique. And integrating the results, we will propose the descriptive representation of assembly instructions which are suitable for robot manipulation using deep learning.
 
 \addtolength{\textheight}{-12cm}   




\section*{ACKNOWLEDGMENT}

This work was supported by Institute for Information \& Communications Technology Promotion (IITP) grant funded by Korea government (MSIT) (No.2019-0-01335, Development of AI technology to generate and validate the task plan for assembling furniture in the real and virtual environment by understanding the unstructured multi-modal information from the assembly manual.

\bibliography{bibtex/bib/IEEEabrv.bib,bibtex/bib/references.bib}{}
\bibliographystyle{IEEEtran}

\end{document}